\documentclass[lettersize,journal]{IEEEtran}
\usepackage{amsmath,amsfonts}
\usepackage{algorithmic}
\usepackage{array}
\usepackage[caption=false,font=normalsize,labelfont=sf,textfont=sf]{subfig}
\usepackage{textcomp}
\usepackage{stfloats}
\usepackage{url}

\usepackage{booktabs}
\usepackage{threeparttable}
\usepackage{multicol}
\usepackage{multirow}
\usepackage{verbatim}
\usepackage{graphicx}
\usepackage{amsmath}
\usepackage{hyperref}
\usepackage{setspace}

\def\BibTeX{{\rm B\kern-.05em{\sc i\kern-.025em b}\kern-.08em
		T\kern-.1667em\lower.7ex\hbox{E}\kern-.125emX}}
\usepackage{balance}

\begin{document}

\title{Visible and Near Infrared Image Fusion Based on Texture Information}

\author{
	Guanyu Zhang,Beichen Sun,Yuehan Qi,Yang Liu
	\thanks{{\itshape(Corresponding author: Yang Liu)}}
	\thanks{Guanyu Zhang, Beichen Sun, Yang Liu (e-mail: liu\_yang@jlu.edu.cn) are with College of Instrumentation \& Electrical Engineering, Jilin University, Changchun 130021, China} 
	\thanks{Yuehan Qi is with Department of Art and Science, Queen's University, Canada}
	}

\maketitle

\begin{abstract}
Multi-sensor fusion is widely used in the environment perception system of the autonomous vehicle. It solves the interference caused by environmental changes and makes the whole driving system safer and more reliable. In this paper, a novel visible and near-infrared fusion method based on texture information is proposed to enhance unstructured environmental images. It aims at the problems of artifact, information loss and noise in traditional visible and near infrared image fusion methods. Firstly, the structure information of the visible image (RGB) and the near infrared image (NIR) after texture removal is obtained by relative total variation (RTV) calculation as the base layer of the fused image; secondly, a Bayesian classification model is established to calculate the noise weight and the noise information and the noise information in the visible image is adaptively filtered by joint bilateral filter; finally, the fused image is acquired by color space conversion. The experimental results demonstrate that the proposed algorithm can preserve the spectral characteristics and the unique information of visible and near-infrared images without artifacts and color distortion, and has good robustness as well as preserving the unique texture.
\end{abstract}


\section{Introduction}

\begin{figure*}[htbp]
	\centering
	\includegraphics[scale=0.7]{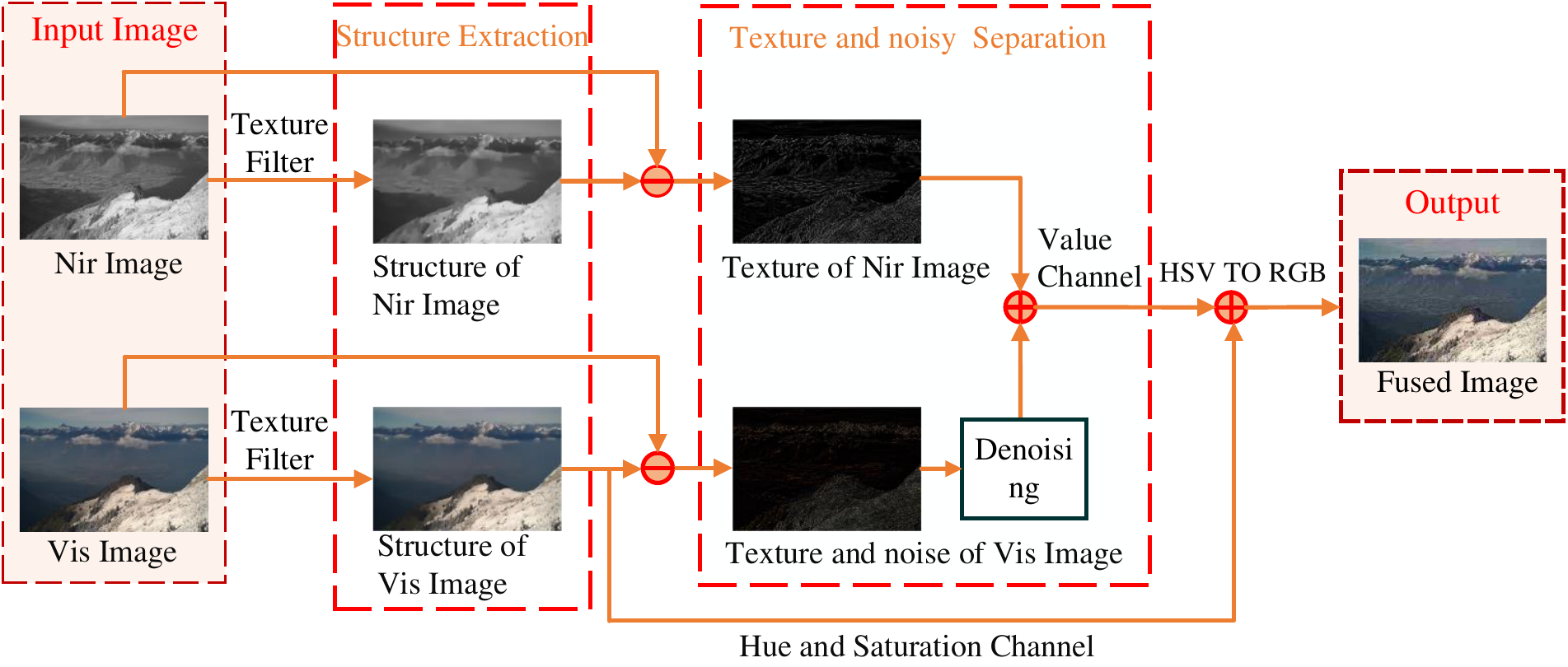}
	\caption{The workflow of our algorithm.}\label{fig:Fig1}
\end{figure*}

At present, autonomous driving technology of unmanned vehicles is the mainstream direction of development in the automotive field, and this gradually developing new technology relies heavily on the performance and collaboration of sensors such as cameras, lidar and IMU, so that the autonomous vehicle can make accurate decisions and control. In the aspect of environmental perception, the existing technology of unmanned vehicle automated driving mainly depends on vision sensors to collect data, and the complexity of the external environment brings serious challenges for visual acquisition tasks. To better adapt to the weather and the effects of illumination change and other environmental mutations, multi-sensor fusion technology more and more reflects the dominant value in terms of visual perception.
RGB (visible) cameras can capture clear and colorful visible light information, but the intensity of the light can affect the clarity of the captured target and is more susceptible to interference. However, NIR (Near Infrared) cameras are less affected by external conditions and capture more information about the structure and texture of the target. Therefore, the fusion of the images captured by the two light sources can solve the visual interference caused by different lighting conditions and produce images with high robustness and better quality.

Currently, most image fusion algorithms use a multi-scale transform to extract and analyze feature information in the target image and then calculate the weight coefficients between the source images as specific fusion strategies. Multi-scale transformation is usually divided into two categories: pyramid transformation in the spatial domain and frequency domain. The pyramid transformation method has many advantages, such as highlighting the details of pictures\cite{2017Visible} \cite{2017Infrared}, and keeping specific scale information. Methods in the frequency domain include wavelet transform\cite{1989A}, dual-tree discrete wavelet transform (DTCWT)\cite{2017Airborne}, nonsubsampled contourlet transform(NSCT)\cite{2013Image}, etc. After completing the multi-scale transformation, it is necessary to reconstruct the information of each scale with specific fusion strategies, adopted the max absolute rule to fuse the low-frequency information\cite{2012The}, and the weighted average underlying information fusion technology based on visual saliency map\cite{2017Infrared}.

However, the multi-scale transformation method, which decomposes the image into high-frequency information and low-frequency information, fails to retain the edge structure of the image as essential information, which will be ignored in the scale transformation. As a result, the intensity of the pixel at the image contour varies greatly, and artifacts appear at the edge of the fused image, which reduces the robustness of the image. Meanwhile, specifying specific fusion strategy is an essential part of image quality. For heterogeneous image fusion, there are differences between the images captured by the two cameras, which will make it impossible to obtain the appropriate weight coefficient, resulting in color distortion and loss of details in the fused image. Therefore, every special structure of the image must be considered when formulating the fusion strategies, otherwise it is difficult to reflect the complementary and cooperative features of image fusion.

The existing NIR and RGB image fusion methods are eliminated the impact of different lighting conditions on the image. Li et al.\cite{9025201} proposed a method to preserve the spectral characteristics based on the reflection and diffusion transport models. The proposed model can compensate for the loss of visible spectral details caused by light scattering. This algorithm can improve the quality of image visibility and avoid image distortion. However, it causes partial loss of NIR texture information. Mohamed Awad et al.\cite{8918077} adopted the method of extracting spatial details of NIR image and adaptively transferring available details into RGB image to enhance fusion image. This method can retain the spectral characteristics of the image and enhance the structural information of the image. However, artefacts appear at the edges of the image. Shibata et al.\cite{2016Versatile} proposed a fusion method based on Markov Random Field (MRF) energy learning to reduce artefacts caused by geometry and illumination inconsistency, but color distortion occurred in the fusion image.Under low light conditions, there is noise in the images taken by the visible camera cause of insufficient light, but the near-infrared camera is not easily affected. Yan et al.\cite{6751301} presented a system effective for cross-field joint image restoration, which can effectively preserve details and edges. Zhou et al.\cite{2020Scale} and Guo et al.\cite{5652900} both use the weighted least squares (WLS) multispectral fusion method to deal with noisy images under low illumination, which have achieved better results in noise reduction, detail transfer and color reproduction rather than keeping the texture details. Mei et al.\cite{9412411} established a convolutional neural network including three sub-networks: DenoisingNet, EnhancingNet and FusionNet to eliminate the noise of noisy pictures under low illumination, but a huge challenge is to obtain a suited and adaptive data set.

In summary, although the current algorithms have made progress, these algorithms do not maintain the spectral characteristics of the visible image, the fused image appears color distortion, and some algorithms will appear artefacts at the edge of the image. In the case of insufficient illumination, it may cause noise in the visible image, so it is essential to remove the influence of noise on the fusion image as well as improve the overall robustness and visual effect of the image.

To solve these problems, this paper proposes a novel and effective fusion framework based on image texture information extraction, automatically fusing NIR and RGB images with Gaussian noise and noiseless environments, and retaining information specific to each light source. Section II introduces a texture filter based on relative total variance and Joint bilateral filter. In Section III, the workflow of the algorithms in this paper is presented. In Section IV, the performance of the algorithms in this paper is evaluated both subjectively and objectively. Finally, Section V concludes the paper.

\section{Edge preserve filter}
	\subsection{Texture filter}
	Texture filter is a filter that retains the main structure of the image edge by smoothing texture, which makes subsequent image processing easier. The texture filter based on relative total variation (RTV) believes that the main edge of local window has a more similar direction gradient than the complex texture part\cite{2012Structure}. A new windowing variant form is adopted to effectively extract the main structure of the image. To highlight the contrast between texture and structure in the prominent area, a more powerful regularization term of structure and texture decomposition is formed. This method has good advantages in timeliness, and its objective function can be expressed as:
	
	\begin{equation}
		\arg \min _{S} \sum_{p}\left(S_{p}-I_{p}\right)^{2}+\lambda \cdot\left(\frac{\mathcal{D}_{x}(p)}{\mathcal{L}_{x}(p)+\varepsilon}+\frac{\mathcal{D}_{y}(p)}{\mathcal{L}_{y}(p)+\varepsilon}\right)
	\end{equation}

	Where $ S_{p} $ is the output image and $ I_{p} $ is the input image, and $\lambda$ is the coefficient representing the smoothing weight, $\varepsilon$ is a very small quantity to prevent the denominator of the formula from being 0.$ D_{x}(p $) and $ D_{y}(p) $ represent the total change of the window of pixel $ p $ in the $ x $ and $ y $ directions, and calculate the absolute spatial difference within the window. $ L(p) $ represents the overall spatial change of pixel $ p $ in $ x $ and $ y $ directions. ($ D_{x}(p) $ / $ L_{x}(p) $ + $\varepsilon$) + ($ D_{y}(p) $ / $ L_{y}(p) $ + $\varepsilon$) used to remove texture effects from images. Equation (1) is written in matrix form as follows:

	\begin{equation}
	\!\left(\!v_{S}\!\!-\!\!v_{I}\!\right)^{\!T\!}\!\!\left(\!v_{S}\!\!-\!\!v_{I}\!\right)\!\!+\!\left(\!v_{S}^{T}\! C_{x}^{T} U_{x} W_{x} C_{x} v_{S}\!+\!v_{S}^{T} C_{y}^{T} U_{y} W_{y} C_{y} \!v_{S}\!\right)
	\end{equation}		

	Where $ v_{S} $ and $ v_{I} $ are vector representations of $ S $ and $ I $ respectively, and $ C_{x} $ and $ C_{y} $ are Toplitz matrices formed by discrete gradient operators from a forward difference. $ U_{x} $, $ U_{y} $, $ W_{x} $, and $ W_{y} $ are all diagonal matrices.
	
	To facilitate calculation, optimizes the solution of Equation (2) at the same time\cite{2012Structure}, and finally expresses it into a linear solution with multiple iterations, which can be expressed as follows:
	
	\begin{equation}
		\left(\mathbf{1}+\lambda L^{t}\right) \cdot v_{S}^{t+1}=v_{I}
	\end{equation}

	Where \textbf{1} is an identity matrix, $ L_{t} $=$ C_{x}^{T} $$ U_{x}^{T} $$ W_{x}^{T} $$ C_{x} $+
	$ C_{y}^{T} $$ U_{y}^{T} $$ W_{y}^{T} $$ C_{y} $is the weight matrix of image vs, and(1+$\lambda$$ L_{t} $) is the symmetric positive definite Laplacian matrix. This method makes the solution faster and the value tends to be stable.
	
	\subsection{Joint bilateral filter}
	Joint bilateral filtering (JBLF) is used as a nonlinear filter\cite{2004Digital}, which achieves edge-preserving and noise-reducing smoothing effects.  It adds the guide image on top of the bilateral filter, and the whole smoothing process is made more robust and stable by the intervention of the guide image to get better results. It is defined as follows:
	\begin{equation}
		S_{p}=\frac{1}{k} \sum_{q \in N(p)} f(p, q) g\left(G_{p}, G_{q}\right) I_{q}
	\end{equation}
	where $ S_{p} $, $ G_{p} $, $ I_{p} $ represent the smoothed, guided and input images under $ p $ pixels, respectively. $ p $=($ p_{x} $, $ p_{y} $), $ q $=($ q_{x} $, $ q_{y} $) are the spatial coordinates of the two pixels, and $ f  $($\cdot$) and $ g $ ($\cdot$) are the spatial and color distances between the two pixels.

\section{The Proposed Algorithm}
To better deal with the structural information of the image edge, we use texture filter to separate the image into two parts: the main structure and the texture. This method can better separate the fine texture details from the edge structure. Meanwhile, it retains the detail changes of the edge features without causing the information loss of the detail layer. We choose the filter based on RTV as the texture filter to split structure and texture. Of course, other texture filters also can be selected.

The visible and near infrared image fusion algorithm proposed in this paper mainly includes the following four steps: (1) separation of structure and texture information, (2) calculation of noise weight, (3) noise removal, and (4) image fusion. The  Fig. \ref{fig:Fig1} shows the workflow.
	\subsection{Separation of structure and texture information}
\begin{figure*}[htbp]
		\centering
		\includegraphics[scale=0.75]{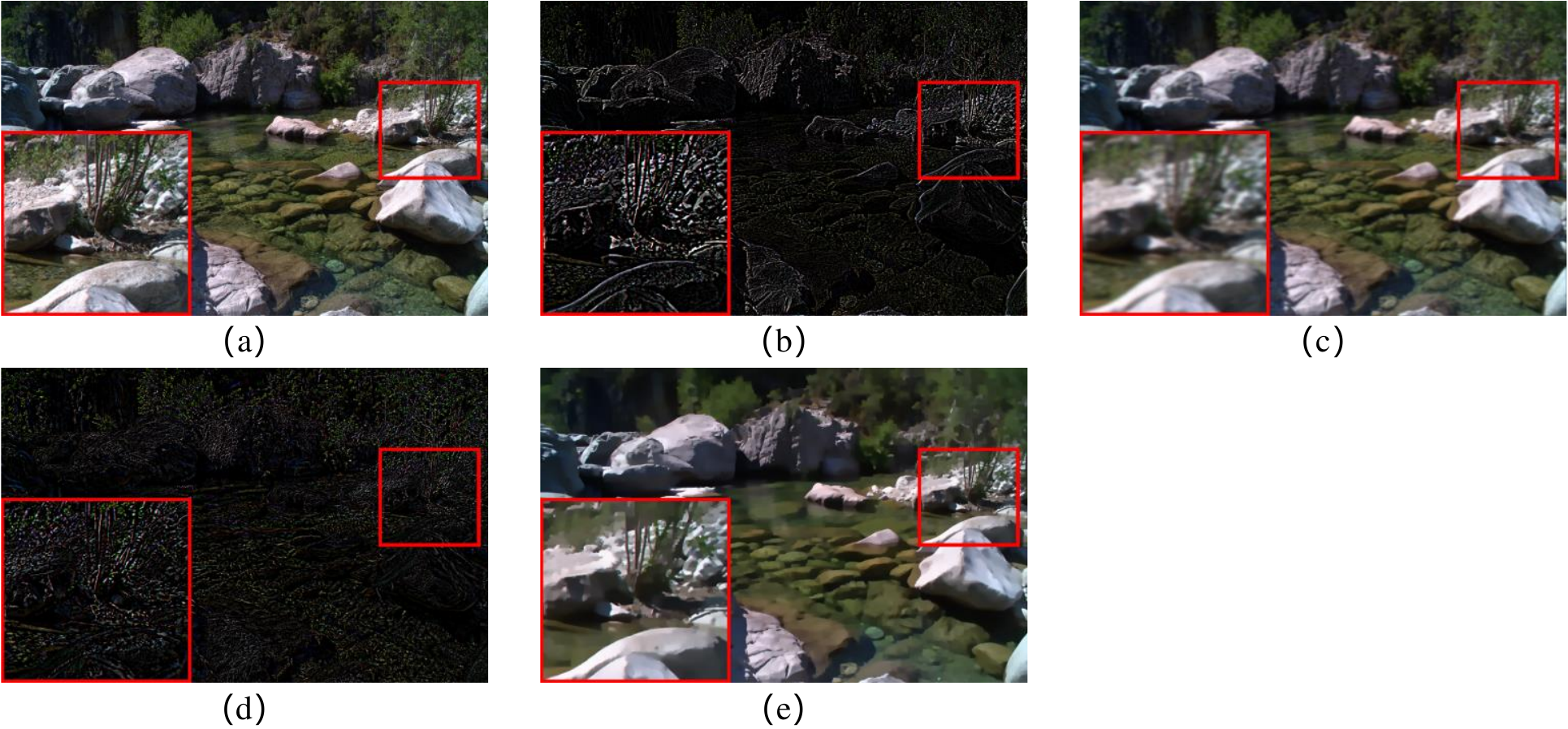}
		\caption{(a) the original picture (b) is the high frequency image obtained by Gaussian filter (c) is the low frequency image (d) is the texture image obtained by RTV (e) is the structure image obtained by RTV}\label{fig:Fig2}
	\end{figure*}

	An image can be decomposed into two parts, the primary structure and the texture, as shown in equation (5):
	\begin{equation}
		I_{C}=T_{C}+S_{C}
	\end{equation}
	Where $ C $$\in$($ vis $, $ nir $) represents the visible and near infrared channels, $ I_{C} $ represents the original image, $ S_{C} $ is the structure image, $ T_{C} $ expresses the final coarse texture image, whose details are smoothed out by RTV texture filtering.
	The significant structure $S_{vis}$ of visible images can be extracted by texture structure separation. $T_{vis}$ is a coarse texture image that needs to be denoised and $ V_{nir} $ is used to calculate the luminance channel of the fused image. Fig. \ref{fig:Fig2} shows the difference between the high-frequency and low-frequency image and the texture image. We can see that the high-frequency information is sharper at the edge of the high-frequency image, while the texture image is more natural, which is the main reason for the artefacts on the edge of the high-frequency information fusion.

	\subsection{calculation of noise weight}
	In low light condition, there will be a lot of noise in the visible image collected by the camera. The coarse texture image $T_{vis}$ needs to be reprocessed to separate the fine texture from the noise to extract the effective information. 
	However, the near infrared image, with high robustness and clear texture, does not contain noise, so its coarse texture information is basically the same as fine texture, and there is no need for secondary separation. For the coarse texture information extracted from the visible image, the noise is randomly distributed in all parts of the image, if a single filtering algorithm is adopted, not only the noise can not be effectively removed, but also the original fine texture will be destroyed, resulting in the loss of information, which is not conducive to the final image fusion.
	
	Zhang et al.\cite{2011Wavelet} mentioned in decomposition and differentiation of noise and texture that different components of images may have diverse local variances. The formula for local variance is as follows:
	\begin{equation}
		P_{N}=\left(\!\frac{1}{|N|}\!\int_{N}\left|f(x)-m_{N}f\right|^{2}\right)^{1 / 2}m_{N} f=\!\frac{1}{N}\! \int_{N} f d x
	\end{equation}
	
	Where $ f $($ x $) represents the local area of the image, $ N $ is the scale information of this part, and $ m_{N} $$ f $ represents the mean value of this area. For different parts of the image, it has the following relationship: $ P_{smooth}>P_{texture}>P_{edge}>P_{noise} $. In other words, the noise part has a weaker variance compared with other regions of the image. We consider the property of noise as a prerequisite for separating fine texture information from noise. However, plentiful texture and noise are mixed in the RGB image, and the local variance image obtained in this way cannot be straightly used as the evaluation standard for judging the robust texture and noise, so the feature factors need to be extracted through reprocessing.

	We use histogram to calculate the distribution of the local variance of each picture, divide the local variance of each image into several parts, solve the maximum value of the first-order gradient and obtain the coordinate information where the value is located as the two feature elements of the image. The two feature elements are put into the naive Bayes model for classification, and the prediction label of each image is obtained, which is marked as a noisy image or noiseless image.
	
	Firstly, we select a 3×3 window to traverse the whole image and calculate the local variance of the brightness channel of the visible image through Formula (6) to obtain the local variance image $ L $.Then we normalize it. Finally, we take the histogram $h$ of $ L $. The expression for acquiring the maximum value of the $h$ first-order gradient and its index is as follows:
	
	\begin{equation}
		f_{1}=\stackrel{}{\max } \nabla h
	\end{equation}
	
	\begin{equation}
		f_{2}=\arg \max \nabla(h)
	\end{equation}
	
	Where we set a training instance $ I $=($ x $, $ c $), $ x $=($ f_{1} $, $ f_{2}$ ) is the characteristic attribute, and $ c $ is the category attribute divided into noisy and noise-free. The probability that instance $ I $ belong to $ c $ can be expressed by Bayes' theorem as:
	\begin{equation}
		P(c \mid x)=\frac{P(c) P(x \mid c)}{P(x)}=\frac{P(c)}{P(x)} \Pi_{i=1}^{d} P\left(x_{i} \mid c\right)
	\end{equation}
	
	Where $ P(c) $ is the prior probability, which can be obtained from the training model. $ P(x| c) $ is the likelihood probability of sample $ I $ relative to class $ c $. $\prod_{i=1}^{d} P(x_{i} |c) $=$ P(x_{1}|c) $×$ P(x_{2} |c) $×...×$ P(x_{d} |c) $ is the product of multiple conditional probabilities in the training data. Finally, we can get that the Bayesian model is:
	
	\begin{equation}
		P(c \mid x)=\operatorname{argmax}_{c} P(c) \Pi_{i=1}^{d} P\left(x_{i} \mid c\right)
	\end{equation}
	
	We selected 225 noisy and non-noisy images as the datasets. To make the individual parameters of the training data set not affected by the whole, it is normalized. Finally, the feature elements of the dataset are put into the Naive Bayes classification model for training, and the final classification model is acquired.
	
	\begin{figure}[htbp]
		\centering
		\includegraphics[scale=0.5]{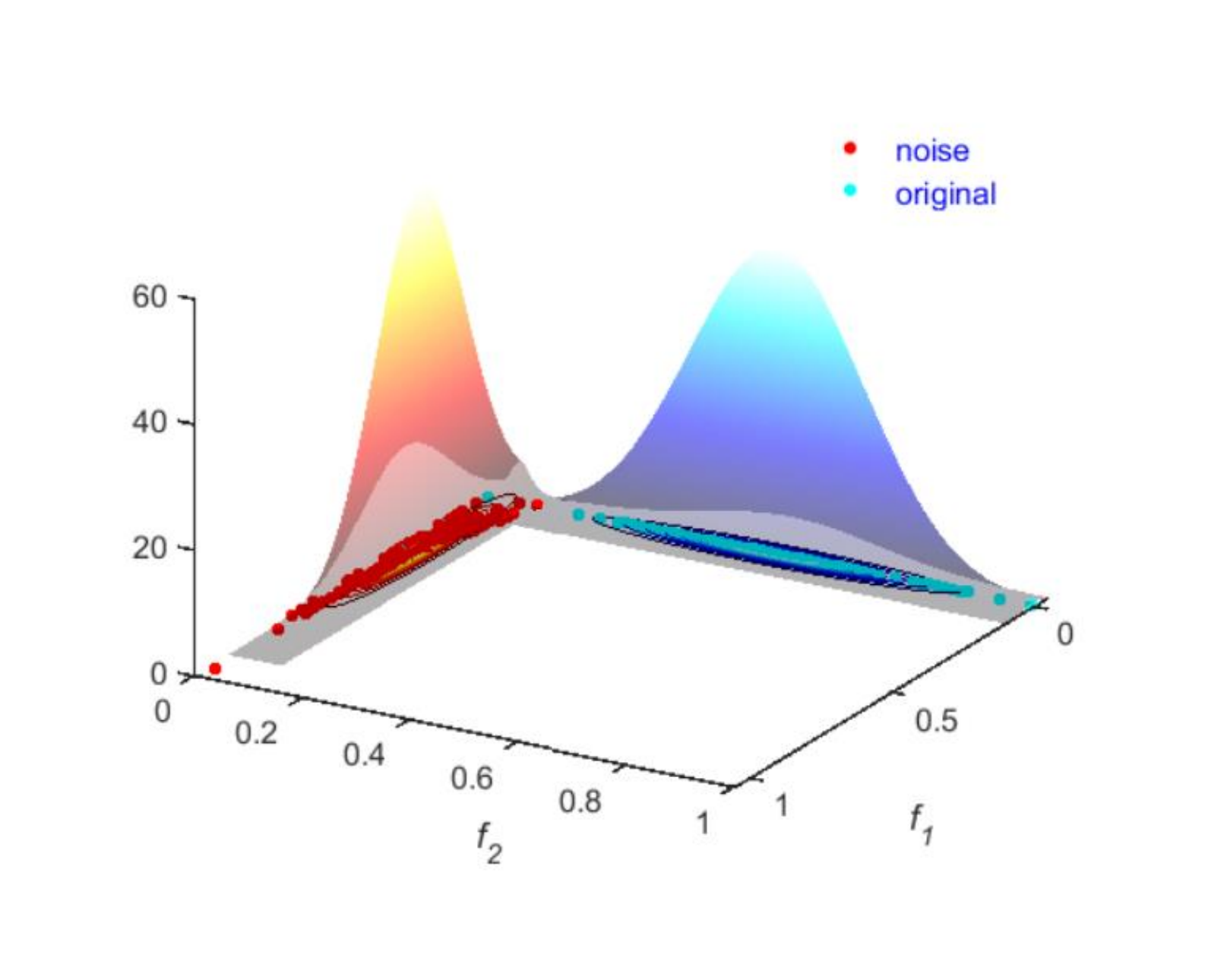}
		\centering
		\caption{Classification results of Bayesian model. (Red) is the probability of Bayesian classification of noisy images; (Blue) is the Bayesian classification probability of original noiseless image.}\label{fig:Fig3}
	\end{figure}
	
	As can be seen from Fig. \ref{fig:Fig3}, noisy and noiseless data are distributed in two categories, so the classification results obtained by the above method can reach the standard of distinguishing noisy and noiseless data.
	\subsection{noise removal}
	
	\begin{figure}[htbp]
		\centering
		\includegraphics[scale=0.45]{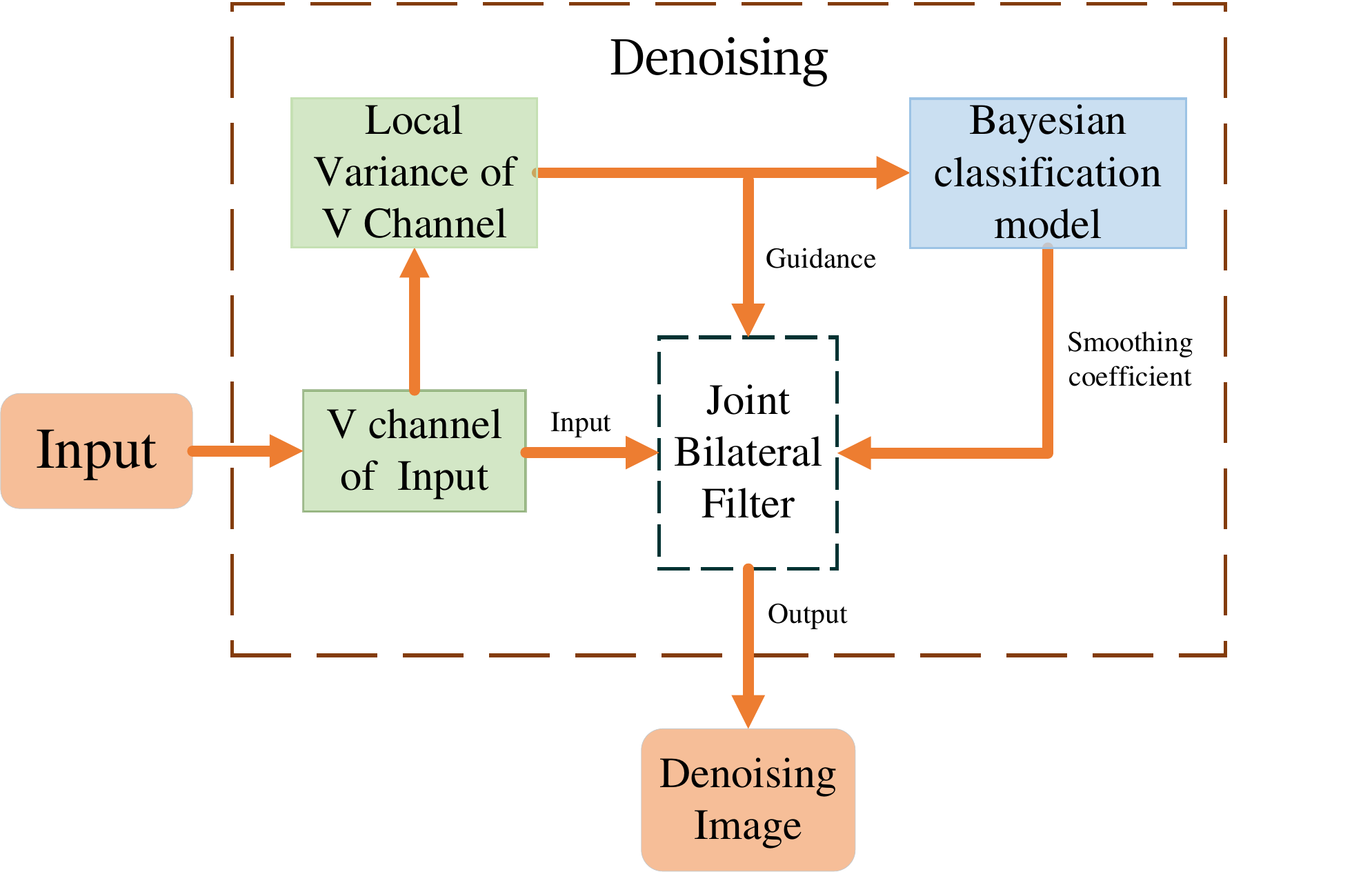}
		\centering
		\caption{Schematic diagram of denoising process.}\label{fig:Fig4}
	\end{figure}

	After the Bayesian model trained in Section 3.2, the following formula is used to calculate the probability $ P_{NOI} $ and $ P_{DNOI} $ of the input image in the noisy and noiseless data.

	\begin{equation}
	\begin{aligned} 
	&P_{T}&=\frac{1}{2 \pi \sigma_{1} \sigma_{2} \sqrt{\!1\!\!-\!\!\rho^{2}\!}} e^{\left(\!-\!\frac{1}{2\left(1\!-\!\rho^{\!2\!}\right)}\left[\frac{\left(x_{\!1\!}\!-\!\mu_{\!1\!}\right)^{\!2\!}}{\sigma_{1}}\!+\!\frac{\left(x_{\!2\!}\!-\!\mu_{\!2\!}\right)^{\!2\!}}{\sigma_{2}}\!-\!\frac{\!2\! \rho\left(x_{\!1\!}\!-\!\mu_{\!1\!}\right)\left(x_{\!2\!}\!-\!\mu_{\!2\!}\right)}{\sigma_{\!1\!} \sigma_{\!2\!}}\right]\right)}
	\end{aligned} 
	\end{equation}

	Where $ T $ $\in$ ($NOI$,$DNOI$) represents two types of noise and noiseless, $ x_{1} $ and $ x_{2} $ respectively represent the normalized($ f_{1} $, $ f_{2} $) of the input image obtained by Section 3.2, $\sigma_{1}$and $\sigma_{2}$ respectively represent the variances of the two feature vectors under T type, $\mu_{1}$ and $\mu_{2}$ represent their mean values, and $\rho$ is the correlation coefficient of the two variables. By default, $\rho$ is 0.5, $\sigma_{1}$ and $\sigma_{2}$, $\mu_{1}$ and $\mu_{2}$ are obtained from the training model.
	
	Aiming to increase the effect of edge-preserving filter smoothing, which is more conducive to noise removal, we add a threshold term to the original JBLF to improve the effectiveness of the algorithm. So change g($\cdot$) of equation (4) to the following form:
	
	\begin{equation}
		\left\{\begin{array}{l}
			1,\left\|G_{p}, G_{q}\right\|_{2}<R \\
			0,\left\|G_{p}, G_{q}\right\|_{2} \geq R
		\end{array}\right.
	\end{equation}

	where $||\cdot||$ denotes the 2-Norm and $ R $ is the threshold value, which we set to 0.2.
	
	Fig. \ref{fig:Fig3} shows the whole image denoising process. We set $ T_{vis} $ derived in Section 3.1 as the input image.The brightness image is obtained from  $ T_{vis} $ by HSV color space transformation
	Then, the local variance L of the brightness image is used as the joint filter to guide the image, and the filter smoothing coefficient is set as $s$. For different noise environments, we constructed a Sigmoid function to adjust the smoothing coefficient $s$ for adaptive filtering.The expression is:
	
	\begin{equation}
		s=\frac{a}{1+e^{-\alpha \cdot k}}+b \quad, \quad k=\frac{P_{DNOI}-P_{NOI}}{\max \left(P_{NOI}, P_{DNOI}\right)}
	\end{equation}

	where $ a $ and $ b $ are the parameters controlling the thresholds.$\alpha$ is a parameter that adjusts the steepness of the curve. $a$ and $b$ are set to 5 and 0.01 respectively $s$ is used as the input parameter of the joint bilateral filter to achieve the effect of voluntarily removing the noise mixed in the noisy image without affecting the noiseless image. The graph below shows the effect of different a value on the smoothing effect. It can be seen that the noise in the image decreases as a increases, but after a certain point, it decreases more and more slowly and consumes more and more time with the value increases.
	
		\begin{figure*}[htbp]
			\centering
			\includegraphics[scale=0.8]{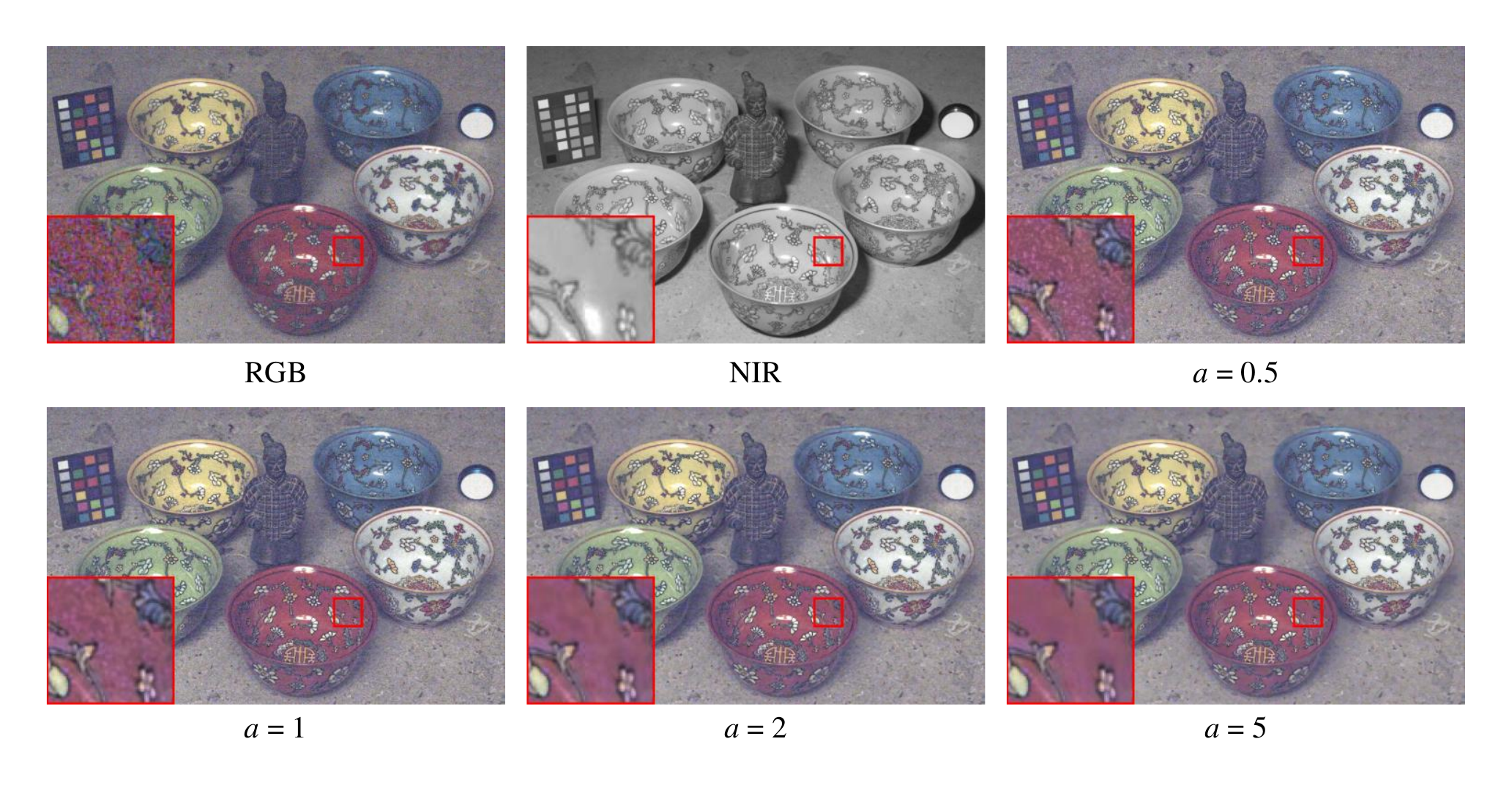}
			\centering
			\caption{Image smoothing results for different parameters $ a $.}\label{fig:Fig5}
		\end{figure*}

	\subsection{Image fusion}
	NIR image is a single-channel image and should not be considered as additional color channels, but rather as channels carrying luminance and spatial information\cite{Cl2008Colouring}. Therefore, we need to convert the visible image from the RGB channel to one of the luminance containing channels such as HSV, YCrCb, or YUV channel and use its luminance channel as part of the fusion with the NIR, so that the color of the fused image can be closer to the color of the source image. We have chosen the V 
	channel in HSV to fuse the NIR information to effectively preserve the NIR information and to prevent color distortion. We use $S_{vis}$ obtained from Section 3.1 to get its tones,
	contrast,and brightness images $H_{rgb}$, $S_{rgb}$, $V_{rgb}$. The fusion brightness image is obtained by the following formula.
	
	\begin{equation}
	F_{V}=V_{rgb}+JBLF\left(T_{vis}, s\right)+T_{nir}
	\end{equation}

	Where $T_{nir}$ is the near-infrared image texture obtained by formula (5) and $JBLF(T_{vis}, s)$ is the result of Section 3.3 combined with bilateral filtering. Finally, we convert the fused brightness image $F_{V}$, $H_{rgb}$ and $S_{rgb}$ into RGB image as the final fusion image.
	
\section{Experimental Results}
In the experiments, we use RGB and NIR datasets with different scene classes\cite{5995637}to verify the merits of the proposed fusion algorithm. And we divided the experimental part into two categories, noiseless and noisy environments. We perform subjective and objective quality comparisons for each set.

To evaluate the performance of the proposed algorithm, a series of contrast experiments are conducted with the state-of-the-art RGB and NIR image fusion methods, including Image Restoration Via Scale Map(VSM)\cite{6751301}, Spectrum Characteristics Preservation(SCP)\cite{9025201}, adaptive and fast image en-hancement(LC)\cite{8918077}, Guided Filter for Fusion(GFF)\cite{6423909}, and fusion using Laplacian-Gaussian Pyramid Decomposition(LGPD)\cite{2017Visible}. To demonstrate that the algorithm can achieve good results in both noisy and noise-free environments, we test it in the noise-free environment Fig. \ref{fig:Fig6}-Fig. \ref{fig:Fig8} and the noisy environment Fig. \ref{fig:Fig9}-Fig. \ref{fig:Fig11}, where the noisy images are obtained by adding Gaussian noise to the noisy images.
	\subsection{Subjective Quality Comparison}
\begin{figure*}[htbp]
		\centering
		\includegraphics[scale=0.7]{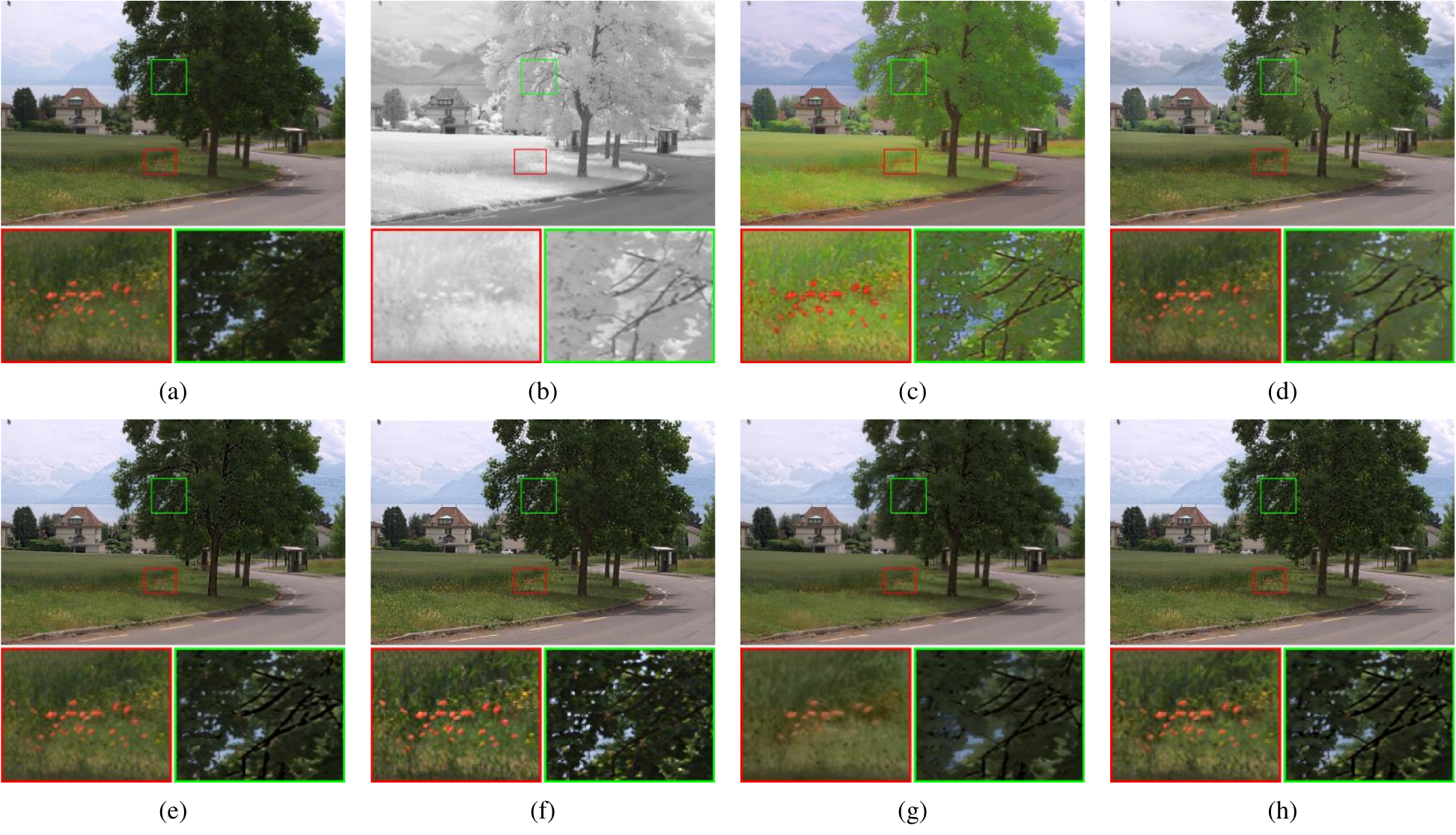}
		\centering
		\caption{Qualitative comparison of different fusion algorithms in image Tree (a) Visible image (b) Near infrared image (c) LGPD\cite{2017Visible} (d) GFF\cite{6423909} (e) LC\cite{8918077} (f) SCP\cite{9025201} (g) VSM\cite{6751301} (h) Our.}\label{fig:Fig6}
	\end{figure*}

\begin{figure*}[htbp]
		\centering
		\includegraphics[scale=0.7]{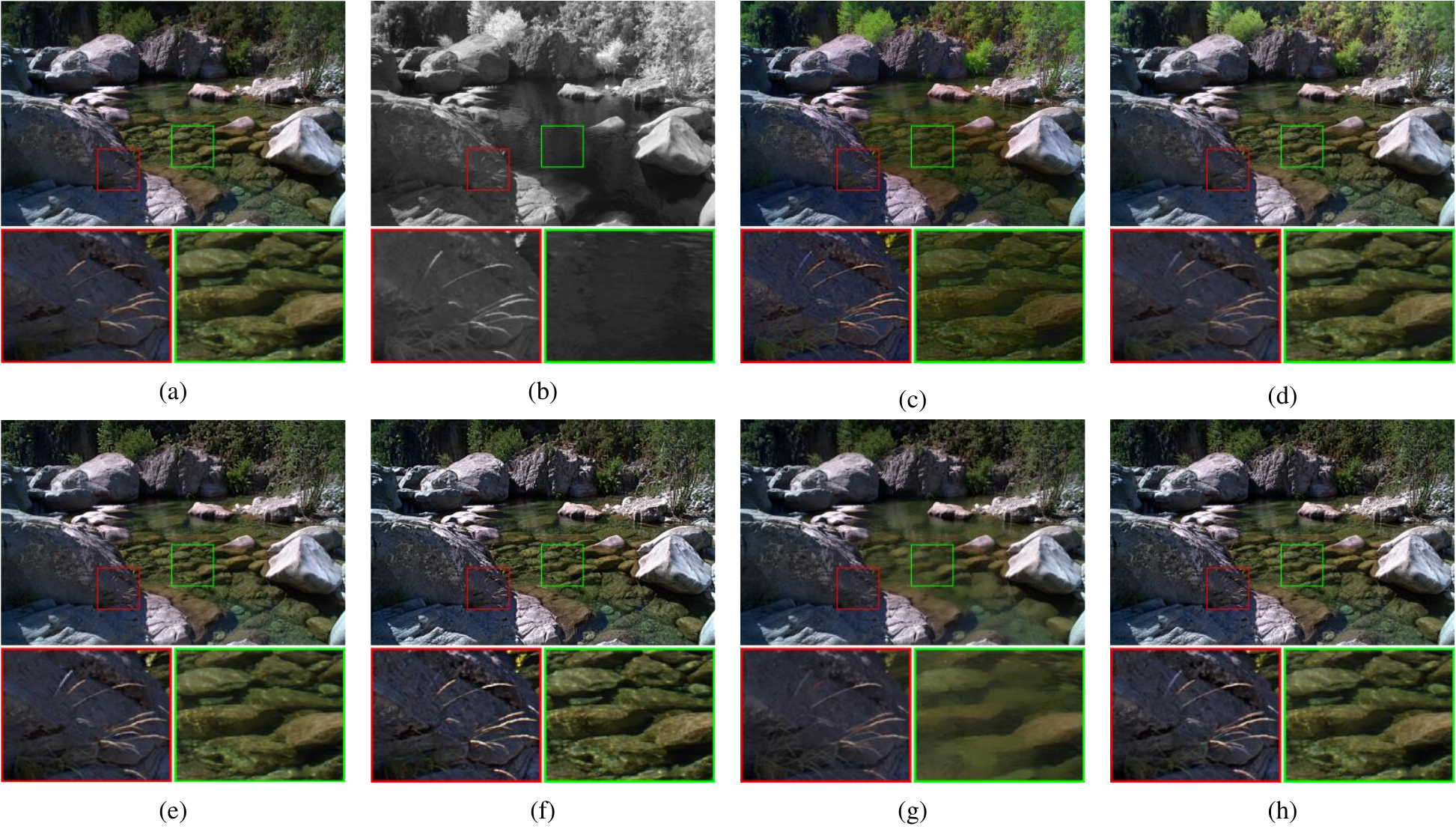}
		\centering
		\caption{Qualitative comparison of different fusion algorithms in image Water (a) Visible image (b) Near infrared image (c) LGPD\cite{2017Visible} (d) GFF\cite{6423909} (e) LC\cite{8918077} (f) SCP\cite{9025201} (g) VSM\cite{6751301} (h) Our.}\label{fig:Fig7}
	\end{figure*}

\begin{figure*}[htbp]
		\centering
		\includegraphics[scale=0.7]{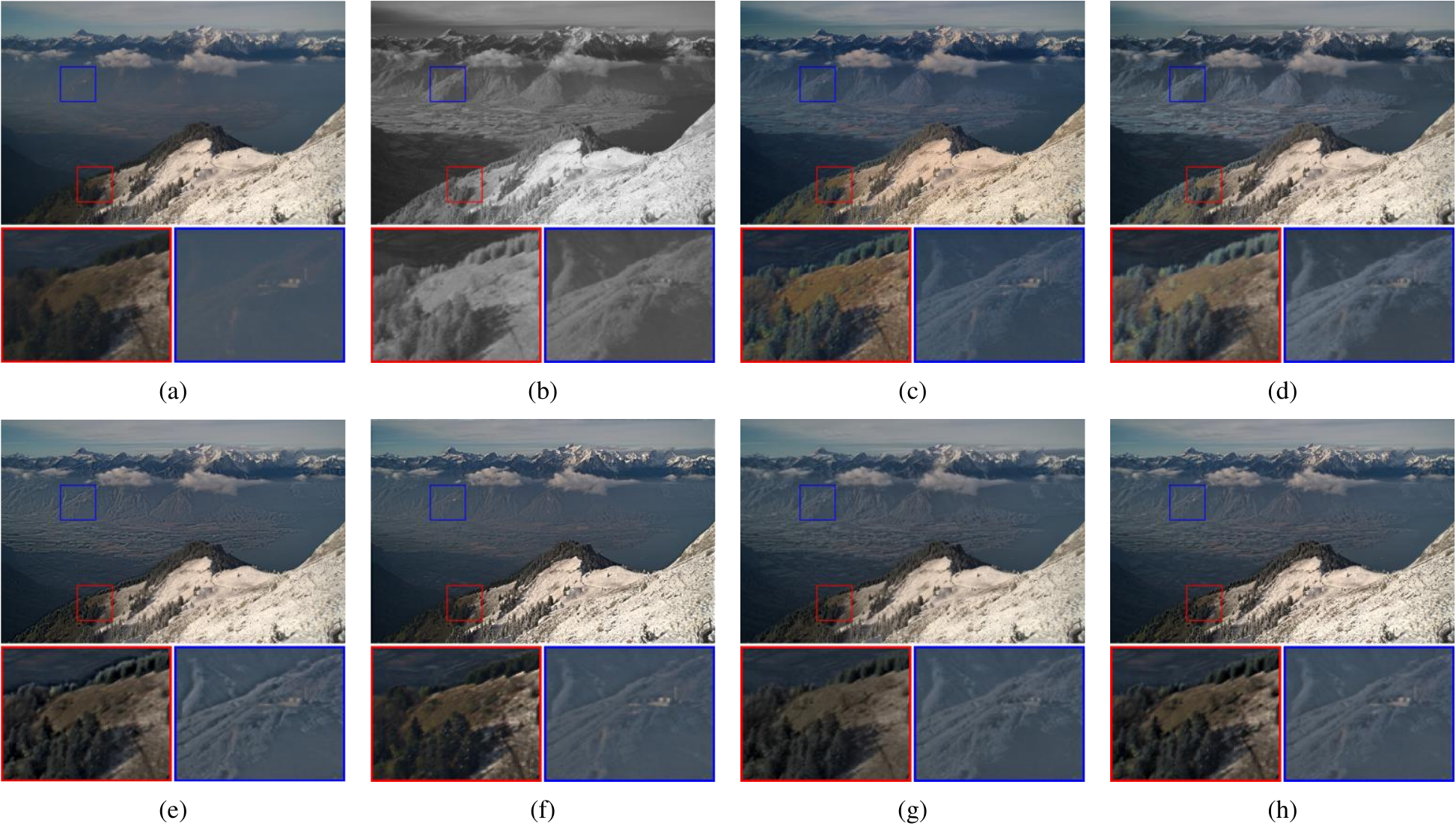}
		\centering
		\caption{Qualitative comparison of different fusion algorithms in image Mountain (a) Visible image (b) Near infrared image (c) LGPD\cite{2017Visible} (d) GFF\cite{6423909} (e) LC\cite{8918077} (f) SCP\cite{9025201} (g) VSM\cite{6751301} (h) Our.}\label{fig:Fig8}
	\end{figure*}

	It can be seen from Fig. \ref{fig:Fig6} to Fig. \ref{fig:Fig8} that LGPD and GFF fusion images are brighter as a whole, and their colors have serious deviations, which do not maintain the spectral characteristics of the original visible image, resulting in less natural color distortion. In Fig. \ref{fig:Fig6}, SCP can keep color changes, but the texture of the branch in the NIR image is not prominent. Overall changes in detail are not highlighted in the fused image. VSM lost part of visible light texture information. In Fig. \ref{fig:Fig6}, there is smoothness in the grass part, which makes the details not obvious. In Fig. \ref{fig:Fig7}, the water part fails to maintain the original rich texture information of visible light, which makes the underwater area unreal, and the tower in the blue box in Fig. \ref{fig:Fig8} is also blurred. Artefacts have appeared at the edges of LC's image in Fig. \ref{fig:Fig8}
	
\begin{figure*}[htbp]
		\centering
		\includegraphics[scale=0.7]{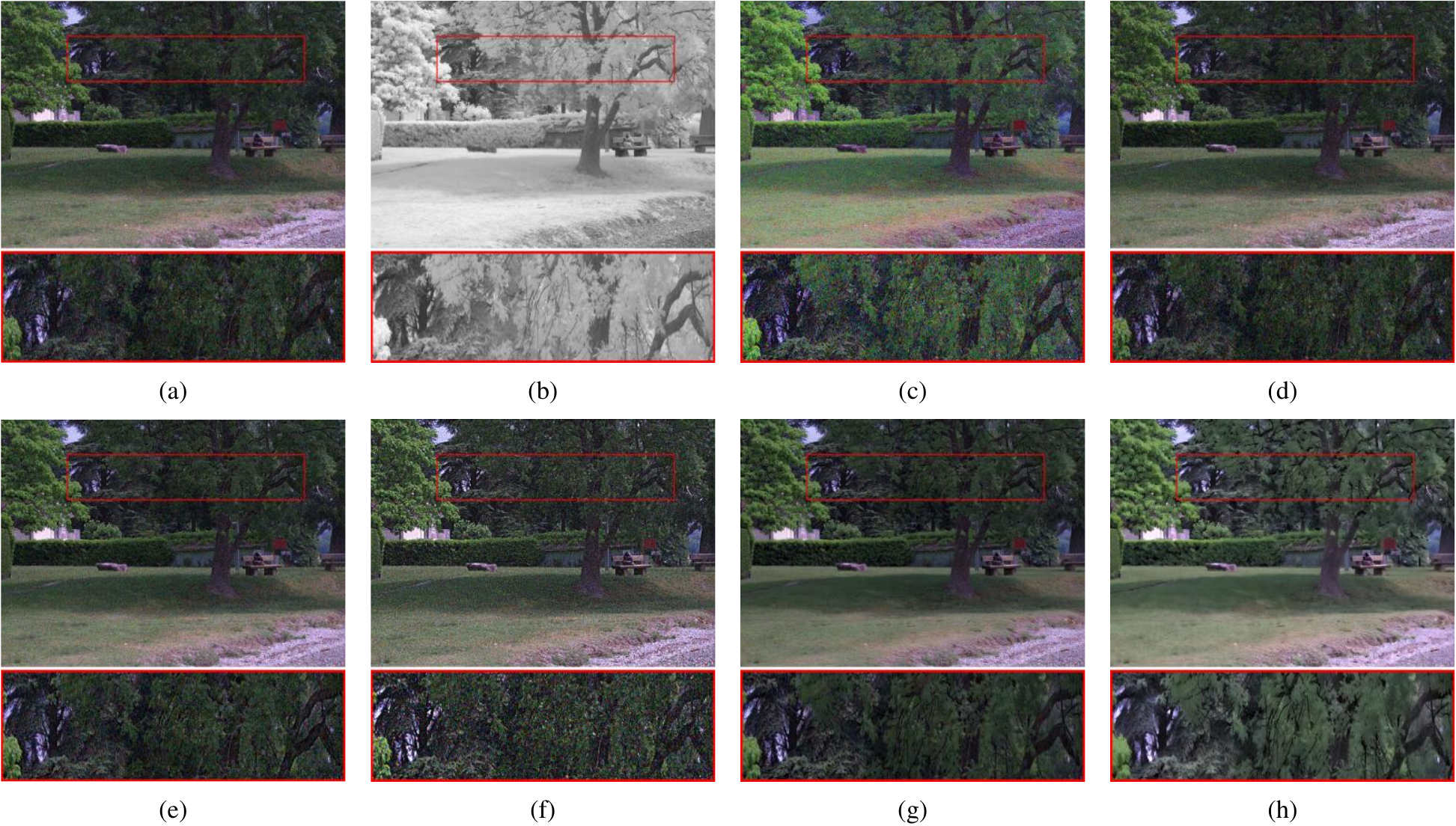}
		\centering
		\caption{Qualitative comparison of different fusion algorithms in image Forest (a) Visible image (b) Near infrared image (c) LGPD\cite{2017Visible} (d) GFF\cite{6423909} (e) LC\cite{8918077} (f) SCP\cite{9025201} (g) VSM\cite{6751301} (h) Our.}\label{fig:Fig9}
	\end{figure*}

\begin{figure*}[htbp]
		\centering
		\includegraphics[scale=0.7]{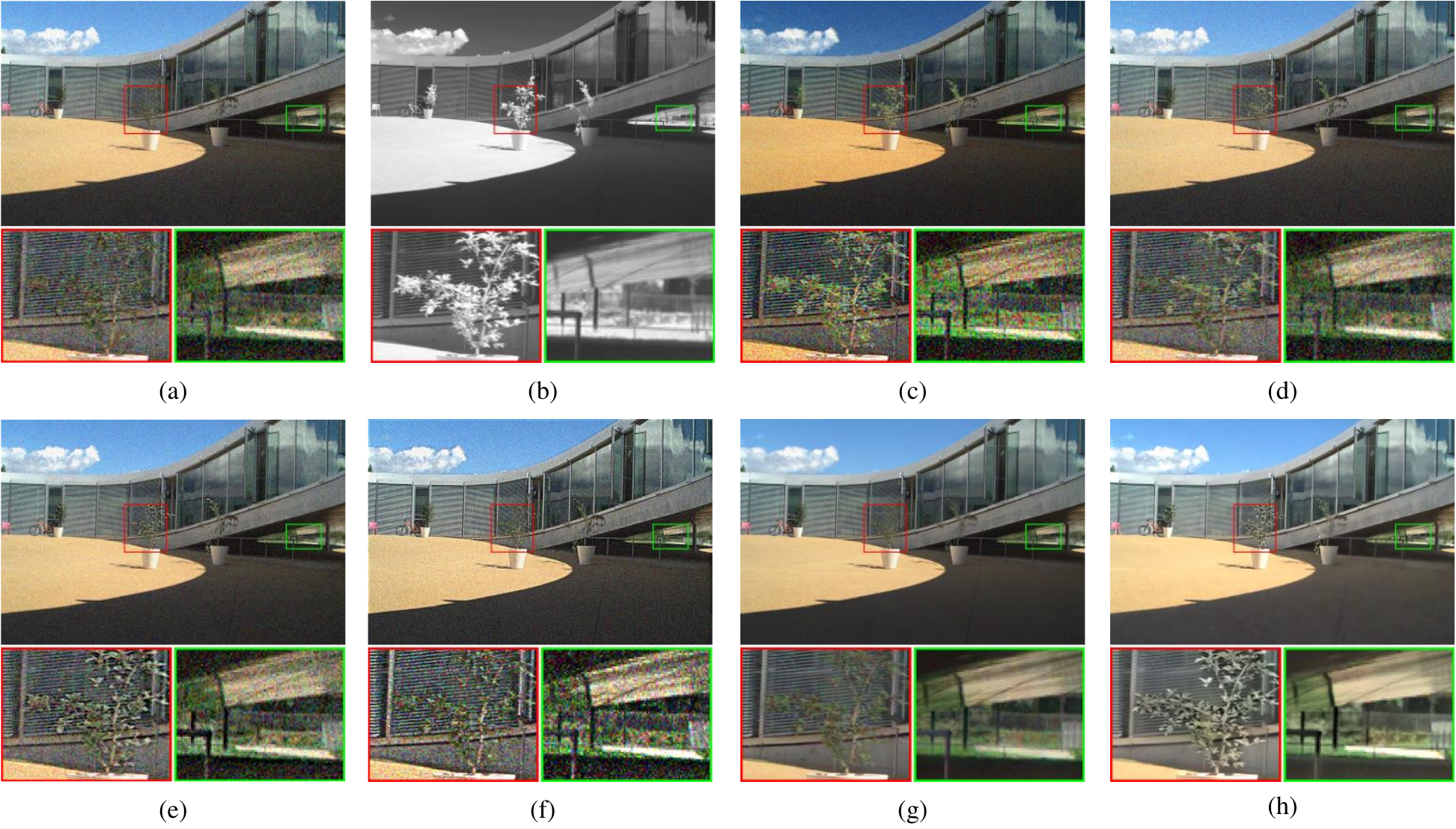}
		\centering
		\caption{Qualitative comparison of different fusion algorithms in image Urban (a) Visible image (b) Near infrared image (c) LGPD\cite{2017Visible} (d) GFF\cite{6423909} (e) LC\cite{8918077} (f) SCP\cite{9025201} (g) VSM\cite{6751301} (h) Our.}\label{fig:Fig10}
	\end{figure*}

\begin{figure*}[htbp]
		\centering
		\includegraphics[scale=0.7]{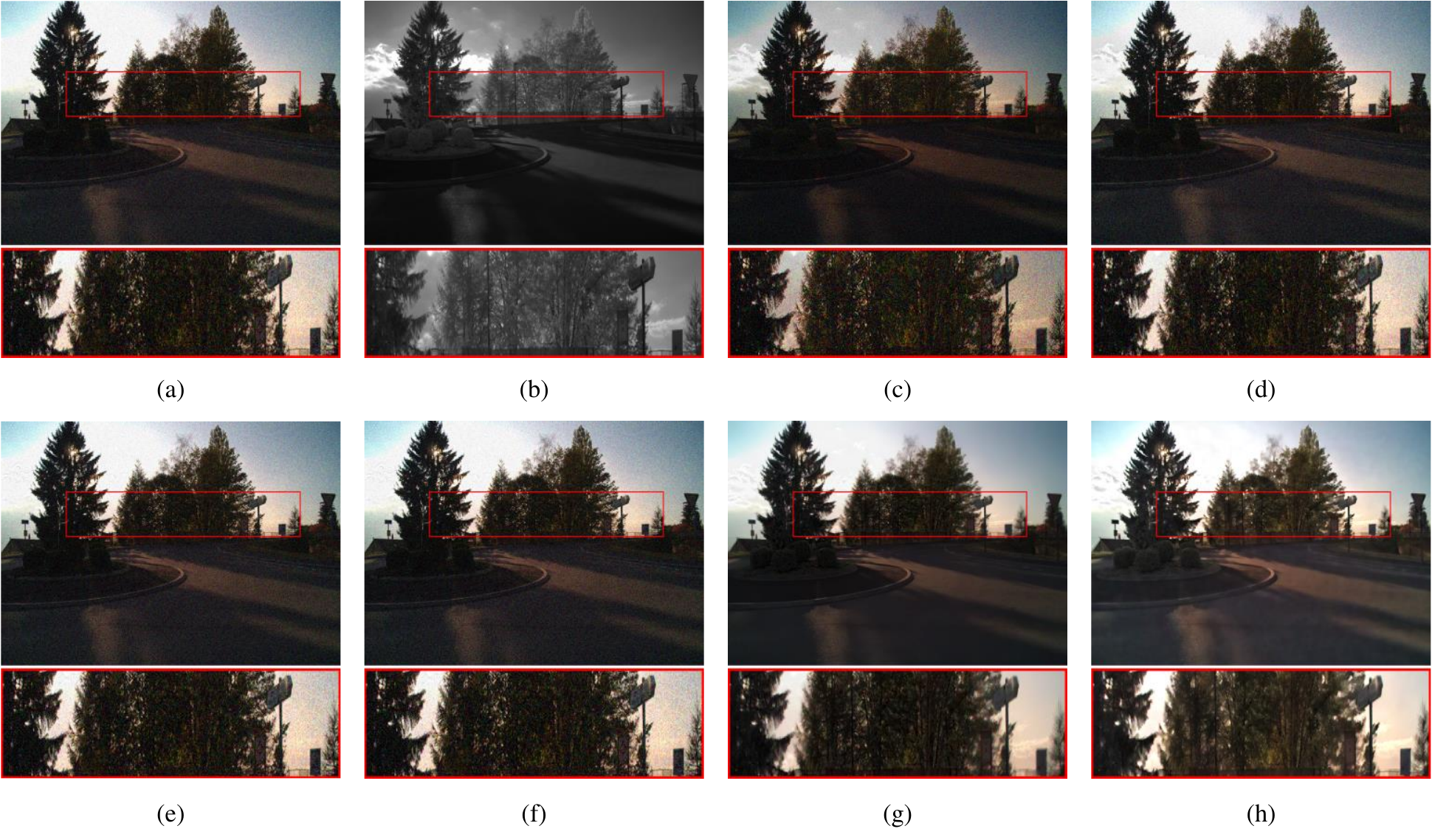}
		\centering
		\caption{Qualitative comparison of different fusion algorithms in image Street (a) Visible image (b) Near infrared image (c) LGPD\cite{2017Visible} (d) GFF\cite{6423909} (e) LC\cite{8918077} (f) SCP\cite{9025201} (g) VSM\cite{6751301} (h) Our.}\label{fig:Fig11}
	\end{figure*}

	In the case of noise, LGPD, GFF, SCP and LC in Fig. \ref{fig:Fig9}-Fig. \ref{fig:Fig11} failed to entirely remove the noise in the visible image, making the final fusion image polluted by noise. Although VSM can effectively remove noise in Fig. \ref{fig:Fig9}-Fig. \ref{fig:Fig11}, the red box part of Fig. \ref{fig:Fig9} is hard to stick out the details of the near-infrared image. In Fig. \ref{fig:Fig10}, the edge structure of the object in the red box is fuzzy and it is hard to highlight the main structure of object.
	
	\subsection{Objective Comparison}
	To objectively compare the image quality of various methods, review\cite{2019Infrared} selected three unreferenced evaluation criteria, structural similarity index measure(SSIM)\cite{1284395}, visual information fidelity for fusion(VIFF)\cite{2013A}, and a comprehensive image fusion quality evaluation index called $ Q^{AB/F} $\cite{2000Objective}.
		\subsubsection{Structural similarity index measure(SSIM)}
		
		Wang et al.\cite{1284395} introduced a complementary framework for quality assessment based on structural similarity. Structural similarity index measure(SSIM) defines structure information from the perspective of image composition, which is different from brightness, contrast, and reflects the attributes of object structure in the scene. Its definition is as follows:
		
		\begin{equation}
			\operatorname{SSIM}(x, y)=\frac{\left(2 \mu_{x} \mu_{y}+C_{1}\right)\left(2 \sigma_{x y}+C_{2}\right)}{\left(\mu_{x}^{2}+\mu_{y}^{2}+C_{1}\right)\left(\sigma_{x}^{2}+\sigma_{y}^{2}+C_{2}\right)}
		\end{equation}
	
		Where SSIM($ x $,$ y $) represents the structural similarity of images to $ x $ and $ y $, $\mu_{x}$ is the pixel mean of $ x $, $\mu_{y}$ is the pixel mean of $ y $, $ \sigma_{xy} $ represents the covariance of two images, $\sigma_{x}$ represents the standard deviation of $ x $, $\sigma_{y}$ represents the standard deviation of $ y $, $ C_{1} $ and $ C_{2} $ are constants used to maintain stability. Finally, the structural similarity between all source images and fusion images can be expressed in the following form:
		
		\begin{equation}
			S S I M=\frac{1}{2}\left(S S I M_{R G B, F}+S S I M_{N R, F}\right)
		\end{equation}
	
		where $ SSIM_{RGB,F} $ and $ SSIM_{NIR,F} $ indicate the structural similarity between the visible and NIR and fused images. Structural similarity allows a more intuitive comparison of the structure of the distorted image and the reference image, and has good performance when comparing to reduce noise capabilities, with larger values indicating that the structure of the fused image is more similar to the original.
		
		\subsubsection{Visual information fidelity for fusion(VIFF)}
		
		The visual Information fidelity for fusion(VIFF) measures the visual natural fidelity and information fidelity of fused images\cite{2013A}, and the multi-resolution image fusion metric using Visual Information Fidelity (VIF) is used to objectively assess fusion performance. This process can be achieved in four steps. Firstly, the source image and fusion image are filtered and divided respectively. Next, the visual information of each block is evaluated with or without distortion. Then, the VIFF of each sub-band is calculated. Finally, the total of VIFF is calculated. The partial expressions are as follows:
		
		\begin{equation}
			\operatorname{VIFF}_{k}\left(I_{1}, \!\ldots\! I_{n}, I_{F}\right)\!=\!\frac{\!\sum_{b}\! F V I D_{k, b}\left(I_{1}, \!\ldots\!, I_{n}, I_{F}\right)}{\!\sum_{b}\! F V I N D_{k, b}\left(I_{1}, \!\ldots\!, I_{n}, I_{F}\right)}
		\end{equation}
	
		where $ I_{1}, I_{2}, ..., I_{n} $ denote the hypothetical n source images, and $ I_{F} $ is the fused image. $ k $ and $ b $ are sub-band and block indexes respectively. $ FVID $ is fusion visual information with distortion and $ FVIND $ represents fusion visual information without distortion. A higher VIFF means a more realistic image with less distortion.
		
		\subsubsection{$ Q^{AB/F} $}
		
		$ Q^{AB/F} $ is a method for objectively assessing fusion performance at the pixel level\cite{2000Objective}. The metric reflects the quality of the edge information obtained from the input image fusion and can be used to compare the performance of different image fusion algorithms. It is defined as:

		\begin{equation}
			 \!Q^{\!AB/F\!}\!\\
			 \!=\!\frac{\!\sum_{N}^{\!i=1\!}\! \!\sum_{M}^{\!j=1\!}\! \!Q^{\!AF\!}(i, j)\! 	w^{\!A\!}(i, j)\!+\!Q^{\!BF\!}(i, j) \!w^{\!B\!}(i, j)}{\!\sum_{N}^{i=1}\! \!\sum_{M}^{j=1}\!\left(w^{A}(i, j)\!+\!w^{B}(i, j)\right)}
		\end{equation}
	
		Where $ Q^{AF}(i,j) $=$ Q_{g}^{AF}(i,j) $$ Q_{a}^{AF}(i,j) $,$ Q^{BF}(i,j) $ =$  Q_{g}^{BF}(i,j) $$ Q_{a}^{BF}(i,j) $,\\
		$ Q_{g}^{F}(i,j) $$ Q_{a}^{F}(i,j) $ denotes the intensity and direction values at position $ (i, j) $ respectively. $ w^{A} $ and $ w^{B} $ denote the weights indicating the importance of each source image to the fused image separately. A larger $ Q^{AB/F} $ value indicates the fused image has more visual information.

\begin{table*}[htbp]
	\centering
	\caption{QUANTITATIVE COMPARISON OF NOISELESS DATA\label{tab:table1}}
	\centering
	\begin{threeparttable}
		\begin{tabular}{cccccccc}
			\toprule[1.5pt]
			Image & Method & LGPD &  GFF &  LC  & SCP  & VSM  & OUR\cr
			\midrule
			\multirow{3}{*}{Tree}		 	
			&SSIM&0.5449&0.5733&0.5762&0.5433&0.4747&\textbf{0.5851}\cr
			&VIFF&0.5239&0.5736&0.6008&0.6032&0.5036&\textbf{0.6755}\cr
			&{$ Q^{AB/F} $}&0.4528&0.6003&\textbf{0.6162}&0.4925&0.4706&0.5225\cr
			\midrule
			\multirow{3}{*}{Water}
			&SSIM&0.6876&\textbf{0.7317}&0.7239&0.6477&0.6916&0.7021\cr
			&VIFF&0.7586&0.8390&0.8275&\textbf{0.9185}&0.7361&0.9006\cr
			&{$ Q^{AB/F} $}&0.5486&0.6938&\textbf{0.7026}&0.4964&0.6137&0.5779\cr
			\midrule
			\multirow{3}{*}{Mountain}
			&SSIM&0.6811&\textbf{0.6952}&0.6778&0.6444&0.6504&0.6734\cr
			&VIFF&0.8478&0.9194&0.8587&0.9547&0.7905&\textbf{0.9859}\cr
			&{$ Q^{AB/F} $}&0.576&\textbf{0.7178}&0.6999&0.5081&0.7152&0.5874\cr
			
			\bottomrule[1.5pt]
		\end{tabular}
	\end{threeparttable}
\end{table*}

\begin{table*}[htbp]
	\centering
	\caption{QUANTITATIVE COMPARISON OF NOISE DATA\label{tab:tavle2}}
	\centering
	\begin{threeparttable}
		\begin{tabular}{cccccccc}
			\toprule[1.5pt]
			Image & Method & LGPD &  GFF &  LC  & SCP  & VSM  & OUR\cr
			\midrule
			\multirow{3}{*}{Forest}
			&SSIM&0.1721&0.2064&0.2226&0.1747&0.5247&\textbf{0.5643}\cr
			&VIFF&0.5060&0.3153&0.3616&0.3824&0.5484&\textbf{0.7119}\cr
			&$Q^{AB/F}$&0.1649&0.1966&0.2154&0.1595&0.4680&\textbf{0.5364}\cr
			\midrule
			
			\multirow{3}{*}{Urban}
			&SSIM&0.1803&0.1863&0.1896&0.1692&\textbf{0.6653}&0.6391\cr
			&VIFF&0.6736&0.5950&0.6029&0.7031&0.7329&\textbf{0.7386}\cr
			&$Q^{AB/F}$&0.2933&0.3101&0.3217&0.2597&0.5571&\textbf{0.5937}\cr
			\midrule
			
			\multirow{3}{*}{Street}
			&SSIM&0.1073&0.1041&0.1059&0.0965&\textbf{0.4582}&0.4535\cr
			&VIFF&0.5067&0.6006&0.6061&0.6755&0.6193&\textbf{0.7334}\cr
			&$Q^{AB/F}$&0.2462&0.2877&0.2898&0.2438&0.3721&\textbf{0.4217}\cr
			\bottomrule[1.5pt]
		\end{tabular}
	\end{threeparttable}
\end{table*}

In Table \ref{tab:table1}, our algorithm maintains a high value for VIFF in the noiseless case, because our algorithm is more capable of separating texture details, making the fused image more realistic and maintaining the natural characteristics of the visible light image capably. LC method has better figures than the other methods. LC adds the structural information of the NIR subject to the visible image, adding richer detail while keeping the source image unchanged, so it has higher values than the other methods, but artefacts can appear at the edges of the image. SCP algorithm is a method of transferring spectral variations by modelling the transmission. This algorithm allows better preservation of the spectral characteristics of visible images and reduces artefacts, so the VIFF is relatively high. Although GFF has a high SSIM and $ Q^{AB/F} $, color distortion is already present in the fused image and it is not conducive to human eye observation. In general, the quality of an image needs to be evaluated from both a subjective and objective point of view.

 As for noisy environment in the Table \ref{tab:tavle2}, our algorithm maintains better results. Other methods are not effective in removing noise in the presence of noise and cannot remove the noise pollution caused by visible light except the VSM method.
 The VSM algorithm has good denoising capabilities as it is guiding the visible image through the NIR, however, it does not take into account that the visible image may contain unique information, so it fails to retain the rich textural properties of the noisy case, leaving the fusion information incomplete. Compared to the VSM method, our algorithm achieves better results by separating texture from noise in visible images and retaining their textural properties. Since our algorithm is adaptive, it could be applied in noisy and noise-free environments, and there is no color distortion, so the fused image is more realistic and more natural.

\section{Conclusion}
In view of the visual perception system is susceptible to environmental influence, this paper proposes a fusion algorithm of visible and near infrared to enhance the image. In contrast to the multi-scale decomposition type of image fusion, we classify the image into structure and texture by absorbing the texture information of the two source images and no longer subdividing the detail layers. The images are subsequently categorised into noisy and noiseless terms with a Bayesian model and guided by their due environment using a joint bilateral filter for adaptive fusion.

Our algorithm has the following virtues:

1. Compared to previous algorithms, we are not guiding the visible image directly by NIR image to fusion and denoise. Instead, we maintain the information unique to each of the visible and NIR.

2. We can retain the spectral characteristics of the color image without causing color distortion and creating artefacts at the edges of the image. In addition, the algorithm has efficient and superior performance in both subjective and objective aspects compared to other fusion algorithms.

\bibliographystyle{IEEEtran}
\bibliography{reference}

\end{document}